%% file: main.tex
\documentclass[10pt,twocolumn,letterpaper]{article}

\usepackage[review]{cvpr}      
\input{preamble}

\definecolor{cvprblue}{rgb}{0.21,0.49,0.74}
\usepackage[pagebackref,breaklinks,colorlinks,citecolor=cvprblue]{hyperref}
\usepackage{array}
\usepackage{tabularx, booktabs, multirow}
\usepackage[none]{hyphenat}

\def\paperID{*****} 
\def\confName{CVPR}
\def\confYear{2024}

\title{CuriosAI Submission to the EgoExo4D Proficiency Estimation Challenge 2025}

\author{%
  \parbox{\textwidth}{\centering
    Hayato Tanoue$^{1}$\hspace{1em}
    Hiroki Nishihara$^{1}$\hspace{1em}
    Yuma Suzuki$^{1}$\hspace{1em}
    Takayuki Hori$^{1}$\\[0.3em]
    Hiroki Takushima$^{1}$\hspace{1em}
    Aiswariya Manoj Kumar$^{1}$\hspace{1em}
    Yuki Shibata$^{1}$\hspace{1em}
    Mitsuru Takeda$^{1}$\\[0.3em]
    Fumika Beppu$^{1}$\hspace{1em}
    Zhao Hengwei$^{1}$\hspace{1em}
    Yuto Kanda$^{1}$\hspace{1em}
    Daichi Yamaga$^{1}$\\[0.35em]
    $^{1}$SoftBank Corp.\, AI \& Data Technology Planning Division\\[0.25em]
    \ttfamily\small
    \{hayato.tanoue, hiroki.nishihara, yuma.suzuki, takayuki.hori,\\
    \hspace{0.2em}hiroki.takushima, aiswariya.manojkumar, yuki.shibata04, mitsuru.takeda, \\
    \hspace{0.4em}fumika.beppu, hengwei.zhao, yuto.kanda, daichi.yamaga01\}@g.softbank.co.jp
  }%
}

\begin{document}
\maketitle
\input{sec/0_abstract}
\input{sec/1_intro}
\input{sec/2_task}
\input{sec/3_method}
\input{sec/4_conclusion}

{
    \small
    \bibliographystyle{IEEEtran}
    \bibliography{main}
}


\end{document}

%% file: preamble.tex
\usepackage[dvipsnames]{xcolor}


%% file: sec/0_abstract.tex
\begin{abstract}
This report presents the CuriosAI team's submission to the EgoExo4D Proficiency Estimation Challenge at CVPR 2025. We propose two methods for multi-view skill assessment: (1) a multi-task learning framework using Sapiens-2B that jointly predicts proficiency and scenario labels ($43.6\ \%$ accuracy), and (2) a two-stage pipeline combining zero-shot scenario recognition with view-specific VideoMAE classifiers ($47.8\ \%$ accuracy). The superior performance of the two-stage approach demonstrates the effectiveness of scenario-conditioned modeling for proficiency estimation.
\end{abstract}

%% file: sec/1_intro.tex
\section{Introduction}
\label{sec:intro}

Recent advances in wearable cameras have enabled first-person video understanding with applications in skill assessment, sports analytics, and education. The EgoExo4D dataset~\cite{egoexo} provides synchronized \emph{egocentric} and \emph{exocentric} views (hereafter referred to as the \emph{ego} and \emph{exo}), offering rich contextual information for activity understanding.

The proficiency estimation task---assessing skill level from multi-view videos---presents unique challenges arising from the subjectivity of skill assessment, high motion variability, and coarse proficiency annotations. Models must capture subtle execution differences yet remain robust to viewpoint variations.

We investigate two methods: a multi-task learning framework built on Sapiens-2B~\cite{sapiens} that jointly processes all views, and a two-stage pipeline that decouples scenario recognition from proficiency assessment using Qwen-VLM~\cite{qwen} and VideoMAE~\cite{videomae}. Our experiments demonstrate that scenario-specific modeling with view-aware classifiers ($47.8\ \%$) outperforms the joint approach ($43.6\ \%$), leading to our final challenge submission.

%% file: sec/2_task.tex
\section{Task Description}
\label{sec:task}

The challenge requires predicting the demonstrator’s proficiency level (\emph{Novice}, \emph{Early Expert}, \emph{Intermediate Expert}, \emph{Late Expert}) from synchronized multi-view videos. Each sample contains five RGB streams (one ego and four exo), covering six scenarios: \emph{Dance}, \emph{Rock Climbing}, \emph{Basketball}, \emph{Music}, \emph{Cooking}, and \emph{Soccer}. Performance is evaluated using Top-1 accuracy on the official test split.

%% file: sec/3_method.tex
\section{Method}
\label{sec:method}

We present two modeling strategies: (1) a multi-task learning framework built on Sapiens-2B that jointly processes all views, and (2) a two-stage pipeline consisting of scenario recognition and view-specific proficiency classification. Validation experiments showed the two-stage approach ($47.8\ \%$) outperformed multi-task learning ($43.6\ \%$), leading to our final submission.

\begin{table}[t]
  \centering
  \caption{Challenge leaderboard results\protect\footnotemark}
  \label{tab:test_overview}
  \begin{tabular}{lc}
    \toprule
    Team & Accuracy (\%) \\
    \midrule
    PCIE EgoPose      & 53 \\
    CuriosAI (Ours)   & 49 \\
    EDPE              & 44 \\
    Baseline          & 43 \\
    \bottomrule
  \end{tabular}
\end{table}
\footnotetext{Leaderboard shows scores in decimal form (e.g., 0.53 corresponds to 53 \%).}

\subsection{Method 1: Sapiens-2B-based Multi-Task Learning}
\label{subsec:method_1}
\input{sec/3-1-ApproachA}

\subsection{Method 2: Scenario-Conditioned Two-Stage Pipeline}
\label{subsec:method_2}
\input{sec/3-2-ApproachB}

%% file: sec/3-1-ApproachA.tex

This method uses the 2-billion-parameter Sapiens-2B model as a unified video encoder for joint proficiency and scenario prediction, processing all five camera views simultaneously.

\paragraph{Architecture.}
Given $T{=}8$ frames per clip, Sapiens-2B outputs features $F \in \mathbb{R}^{B \times T \times 2048}$, where $B$ is the batch size and $T$ the number of frames. Temporal mean pooling yields clip representations $\tilde{\mathbf{f}} = \frac{1}{T}\sum_{t=1}^{T} F_{:,t,:}$, which feed two linear heads for proficiency ($W_{\text{prof}} \in \mathbb{R}^{4 \times 2048}$) and scenario ($W_{\text{scen}} \in \mathbb{R}^{6 \times 2048}$) prediction.

\paragraph{Training.}
We sample eight frames via linear interpolation and jointly process all views. The multi-task objective balances both losses with $\alpha = 0.5$:
\begin{align}
\mathcal{L} ={} &
  \underbrace{\alpha\,\mathcal{L}_{\mathrm{CE}}\!\bigl(\mathbf{W}_{\mathrm{prof}}\tilde{f},\, y_{\mathrm{prof}}\bigr)}
              _{\text{proficiency}}
  \nonumber\\[2pt]
  &+
  \underbrace{(1-\alpha)\,\mathcal{L}_{\mathrm{CE}}\!\bigl(\mathbf{W}_{\mathrm{scen}}\tilde{f},\, y_{\mathrm{scen}}\bigr)}
              _{\text{scenario}}.
\label{eq:multitask}
\end{align}

\begin{figure}[b]
  \centering
  \begin{subfigure}[b]{0.49\linewidth}
    \centering
    \includegraphics[width=\linewidth]{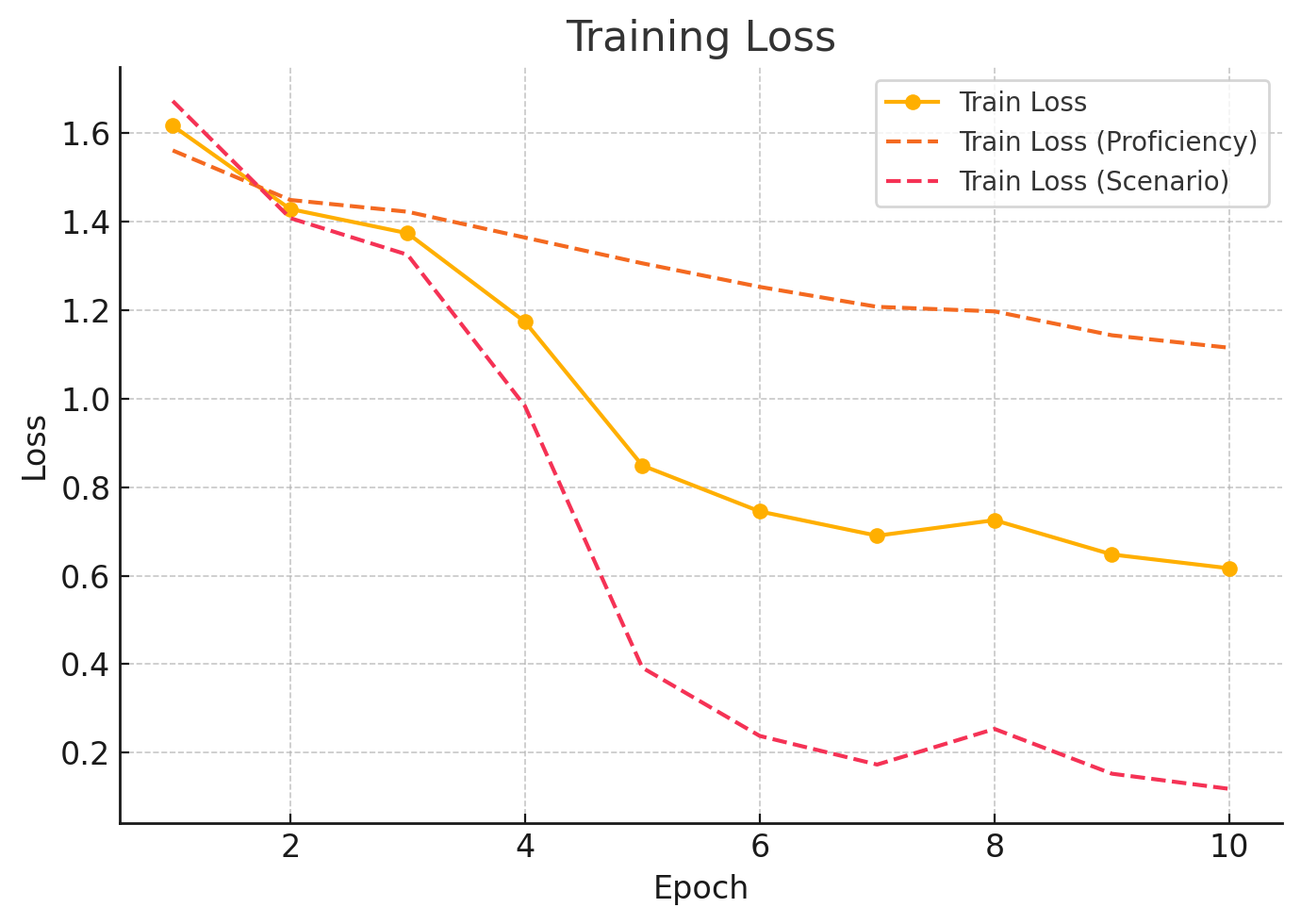}
    \caption{Training loss}
  \end{subfigure}\hfill
  \begin{subfigure}[b]{0.49\linewidth}
    \centering
    \includegraphics[width=\linewidth]{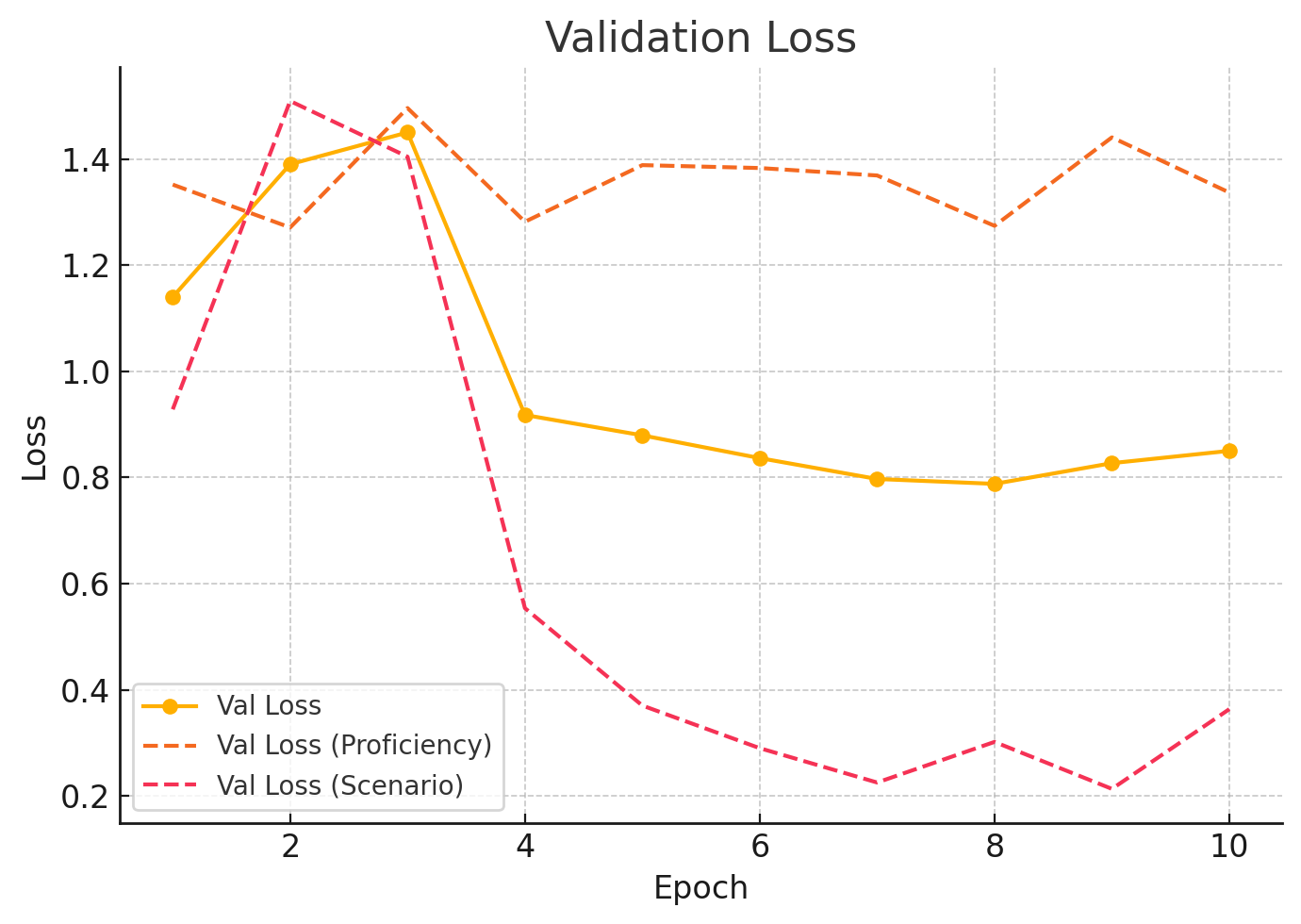}
    \caption{Validation loss}
  \end{subfigure}
  \caption{Training and validation loss for the multi-task model. 
           Solid = total loss, dashed = proficiency sub-loss, dotted = scenario sub-loss.}
  \label{fig:loss_curves}
\end{figure}

\paragraph{Results.}
The model achieves $43.6\ \%$ validation accuracy with significant variation across scenarios (Table~\ref{tab:val_results}). Figure~\ref{fig:loss_curves} shows scenario recognition converges faster than proficiency estimation, suggesting different task complexities in the multi-task objective.

\begin{table}[t]
  \centering
  \caption{Method 1 validation accuracy (\%).}
  \label{tab:val_results}
  \begin{tabular}{lc}
    \toprule
    Scenario & Accuracy (\%) \\
    \midrule
    Basketball     & 39.0 \\
    Cooking        & 30.0 \\
    Dance          & 37.4 \\
    Music          & 65.8 \\
    Rock Climbing  & 45.3 \\
    Soccer         & 68.8 \\
    \midrule
    Overall & \textbf{43.6} \\
    \bottomrule
  \end{tabular}
\end{table}

%% file: sec/3-2-ApproachB.tex

This method addresses the evaluation-only track by training exclusively on the official training split with publicly available pre-trained weights. We employ a two-stage pipeline: first, \textbf{Qwen-2.5-VLM-7B} performs zero-shot scenario recognition, then scenario-specific classifiers estimate demonstrator proficiency conditioned on both the predicted scenario and camera view.

\paragraph{Architecture.}
For scenario detection, we leverage zero-shot Qwen-2.5-VLM-7B without fine-tuning. For proficiency classification, we fine-tune VideoMAE-v2-Huge end-to-end, adding a single linear layer ($1024 \rightarrow 4$) that maps the [CLS] token to four proficiency logits. To capture scenario- and view-specific patterns, we instantiate 30 specialized models—one for each combination of 6 scenarios and 5 camera views.

\paragraph{Training Details.}
Input clips comprise $T{=}16$ uniformly sampled frames, resized to $224 \times 224$ via shorter-side resizing and centre-cropping, then normalised using VideoMAE statistics. We optimize all parameters with AdamW (learning rate $10^{-4}$, weight decay $10^{-2}$) for 20 epochs using a batch size of 8 clips per GPU.

\paragraph{Inference.}
Our inference pipeline proceeds as follows: (1) obtain scenario $s$ via zero-shot Qwen-VLM prediction; (2) for each view $v \in \{\text{ego}, \text{exo1--4}\}$, run the corresponding $(s,v)$-specific classifier to obtain logits $\ell_v \in \mathbb{R}^4$; (3) convert to probabilities via $p_v = \operatorname{softmax}(\ell_v)$; and (4) aggregate predictions using one of three strategies:
\begin{itemize}[nosep,leftmargin=*]
  \item \textbf{Ego only:} $p^{\text{ego}} = p_{\text{ego}}$
  \item \textbf{Exo average:} $p^{\text{exo}} = \tfrac{1}{4}\sum_{i=1}^{4} p_{\text{exo}i}$
  \item \textbf{Combined:} $p^{\text{comb}} = \tfrac{1}{5}\bigl(p_{\text{ego}} + \sum_{i=1}^{4} p_{\text{exo}i}\bigr)$
\end{itemize}

\paragraph{Results.}
Table~\ref{tab:approachB_overall} reports overall validation accuracy, with the five-view fusion achieving $\textbf{47.8\ \%}$---a $4.2\ \%$ absolute improvement over Method 1. This performance gain demonstrates the effectiveness of scenario-specific modeling. The ego-only configuration ($44.3\ \%$) slightly outperforms Method 1's multi-view approach, while exo averaging ($46.9\ \%$) provides substantial gains.

Per-scenario analysis (Table~\ref{tab:approachB_perScenario}) reveals interesting view-dependent patterns. Exo views excel for activities requiring global spatial awareness (Dance: $59.3\ \%$ vs.\ $47.2\ \%$ ego; Music: $68.4\ \%$ vs.\ $57.9\ \%$ ego), while ego views prove superior for spatially constrained tasks (Rock Climbing: $44.0\ \%$ ego vs.\ $34.6\ \%$ exo). Soccer shows viewpoint invariance ($62.5\ \%$ for both), likely due to standardized player movements and consistent field constraints. These results validate our hypothesis that different scenarios benefit from different viewpoint configurations, justifying the view-specific classifier design.

\begin{table}[t]
  \centering
  \caption{Validation Top-1 accuracy (\%) of Method 2.}
  \label{tab:approachB_overall}
  \begin{tabular}{lccc}
    \toprule
          & Ego & Exo (avg) & Combined \\
    \midrule
    Top-1 Acc. & 44.3 & 46.9 & \textbf{47.8} \\
    \bottomrule
  \end{tabular}
\end{table}

\begin{table}[t]
  \centering
  \caption{Per-scenario validation accuracy (\%) of Method 2.}
  \label{tab:approachB_perScenario}
  \begin{tabular}{lccc}
    \toprule
    Scenario        & Ego & Exo & Combined \\
    \midrule
    Basketball      & 34.0 & 40.0 & \textbf{44.0} \\
    Cooking         & 40.0 & \textbf{50.0} & \textbf{50.0} \\
    Dance           & 47.2 & \textbf{59.3} & \textbf{59.3} \\
    Music           & 57.9 & \textbf{68.4} & \textbf{68.4} \\
    Rock Climbing   & \textbf{44.0} & 34.6 & 34.6 \\
    Soccer          & \textbf{62.5} & \textbf{62.5} & \textbf{62.5} \\
    \bottomrule
  \end{tabular}
\end{table}

%% file: sec/4_conclusion.tex
\section{Conclusion}
\label{sec:conclusion}

We presented two methods for the EgoExo4D Proficiency Estimation Challenge at CVPR 2025: multi-task learning with Sapiens-2B and a two-stage pipeline with scenario-conditioned VideoMAE classifiers. Our experiments demonstrate that decoupling scenario recognition from proficiency estimation yields superior performance ($47.8\ \%$ vs. $43.6\ \%$), highlighting the importance of scenario-specific modeling. The analysis reveals distinct view-dependent patterns across activities, with exo views excelling for spatially complex tasks and ego views for constrained activities. These findings suggest that adaptive view selection based on scenario characteristics could further improve proficiency estimation in multi-view settings.